\documentclass[letterpaper, 10 pt, conference]{ieeeconf}  

\IEEEoverridecommandlockouts                              

\usepackage{times}

\usepackage{multicol}
\usepackage[bookmarks=true]{hyperref}
\usepackage{amsmath,amssymb,amsfonts}
\usepackage{booktabs}
\usepackage{caption}
\usepackage{algorithmic}
\usepackage{graphicx}
\usepackage{xcolor}
\usepackage{bbm}
\usepackage{nicefrac}
\usepackage{comment}
\usepackage[font=footnotesize]{caption}

\newcommand\argmin[1]{\underset{#1}{\text{argmin }}}
\newcommand\argmax[1]{\underset{#1}{\text{argmax }}}
\newcommand\p[2]{\frac{\partial #1}{\partial #2}}

\newcommand{\sref}[1]{Sec. \ref{#1}}
\newcommand{\figref}[1]{Fig. \ref{#1}}
\newcommand{\tabref}[1]{Table \ref{#1}}

\title{\LARGE \bf
Safe and Efficient Exploration of Human Models \\During Human-Robot Interaction
}

\author{Ravi Pandya and Changliu Liu
\thanks{Authors are with the Robotics Institute at Carnegie Mellon University, Pittsburgh, PA, USA \tt\small \{rapandya, cliu6\}@andrew.cmu.edu}
}

\begin{document}

\maketitle
\thispagestyle{empty}
\pagestyle{empty}

\begin{abstract}
Many collaborative human-robot tasks require the robot to stay safe and work efficiently around humans. Since the robot can only stay safe with respect to its own model of the human, we want the robot to learn a good model of the human in order to act both safely and efficiently. This paper studies methods that enable a robot to safely explore the space of a human-robot system to improve the robot's model of the human, which will consequently allow the robot to access a larger state space and better work with the human. In particular, we introduce active exploration under the framework of energy-function based safe control, investigate the effect of different active exploration strategies, and finally analyze the effect of safe active exploration on both analytical and neural network human models.
\end{abstract}

\IEEEpeerreviewmaketitle

\section{Introduction}
In order to improve the effectiveness of robots as collaborators for humans, they need to move both safely and efficiently in shared spaces with humans such as in autonomous driving, collaborative manufacturing or social navigation. Since robots should never harm people, safe control is a fundamental system requirement for human-robot interaction and it appears as a hard constraint in real-time control. Recent work has handled this constraint through a safety monitor or safety controller that keeps the robot safe with respect to the human's dynamics \cite{liu2014safeset, fisac2018safeplanning}. Human behavior is often noisy and unpredictable, so the robot should also keep a probabilistic model of the human. Additionally, the robot gets observations of the human online and can use this data to adapt the human model \cite{liu2015safeexploration, fisac2018safeplanning}. 


Much work assumes a ``human-in-isolation'' model  \cite{bajcsy2020robust, ziebart2009planning}, meaning that they do not consider the effect that the robot has on the human. In this work, we make use of the fact that the human will respond to the robot's actions by having the robot actively explore in order to learn a better model of the human. This exploration can be thought of as optimizing for long-term efficiency and is treated as soft constraint rather than a hard constraint like safety for the robot.

In the evaluation, we consider the effects of active exploration in safe control under two kinds of uncertainty: \textit{intrinsic} uncertainty which corresponds to uncertainty in how the human moves in certain areas of the environment and \textit{interactive}  uncertainty which corresponds to uncertainty in how the human reacts to the robot. We additionally consider a data-driven neural network dynamics model where these two sources of uncertainty may be coupled.

The main contributions of this work are to build on top of safe control by 1) introducing active exploration under an energy-function-based safe control framework 2) investigating different strategies for safe exploration and evaluating their efficiency in information collection and 3) investigating the effects of safe exploration on both analytical and data-driven models. Our results suggest that some safe exploration strategies can improve the robot's model of the human and expand the set of safe reachable states, ultimately meaning the robot can improve both the safety and adaptation efficiency of the overall system. 

\begin{figure}[t!]
    \centering
    \includegraphics[width=\columnwidth]{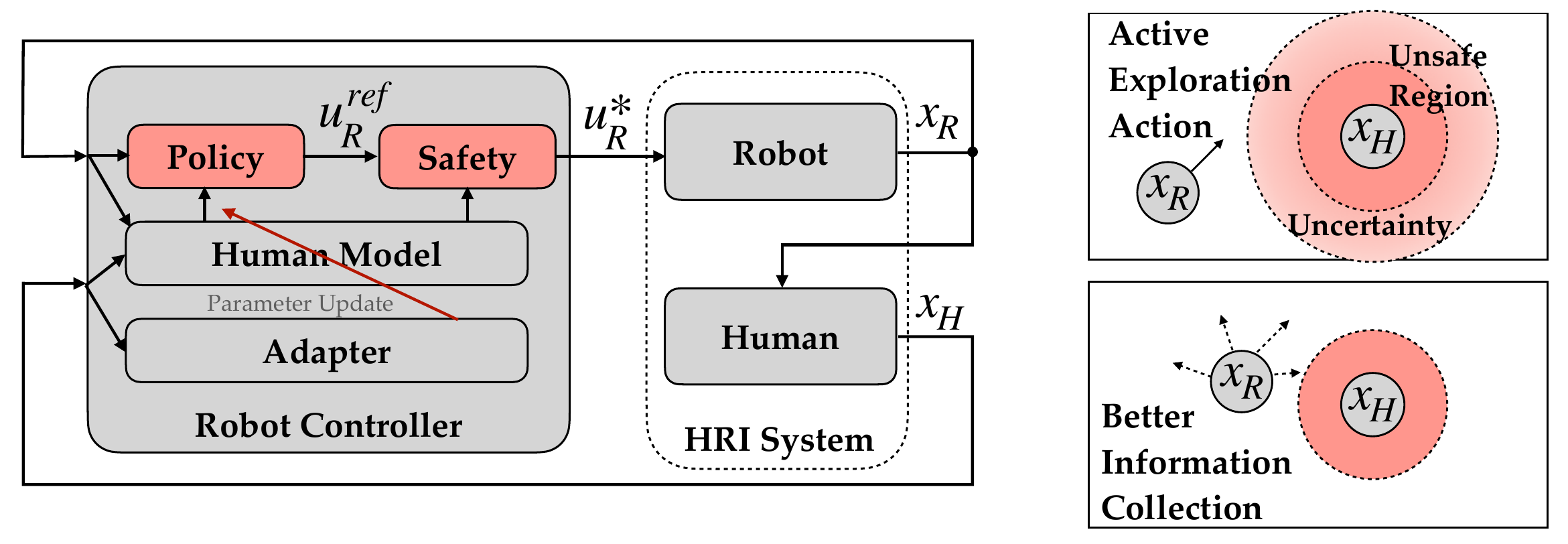}
    \caption{\textbf{Left:} The human-robot system considered in this work, the robot adapts its human model online and is equipped with a safety controller to (probabilistically) guarantee safety. \textbf{Right:} Shows that active exploration can result in reduced uncertainty, hence the robot can access same states with higher safety probability (improved safety) and better collect information around human (improved efficiency).}
    \label{fig:front_figure}
    \vspace{-0.25in}
\end{figure}

\section{Related Work}
\noindent\textbf{Safe control in human-robot interaction:} Since humans and robots will start to work in physical proximity, much work has gone into measuring safety of human-robot systems from collaborative manufacturing to healthcare \cite{michalos2015design, ikuta2003safety}. Others have focused on applying energy-function-based methods \cite{liu2014safeset, ames2019control, wei2019safe} and reachability-based methods \cite{fisac2018safeplanning, bajcsy2019scalable} to mathematically guarantee safety of a robot around humans. In this work, we adopt energy-function-based safe control.

\noindent\textbf{Safe control under uncertainty:} Since human behavior is often noisy and hard to predict exactly, we care about keeping the human-robot system safe even under uncertainty. The general problem of decision making under uncertainty is often modeled as a Partially Observable Markov Decision Process (POMDP). However, they are computationally intractable to solve explicitly \cite{papadimitriou1987complexity}. There is some work which utilizes the exact POMDP solution for decision-making around humans \cite{chen2018trust}, but such approaches do not scale to environments larger than a handful of discrete states and actions. While it is also possible to add hard constraints \cite{isom2008piecewise} or chance constraints \cite{thiebaux2016rao}, these problems are harder than standard POMDPs. 

Another body of work incorporates uncertainty into the safety monitor, such as by using Bayesian inference over the human's intention to inform the safety monitor which states the human is likely to reach \cite{fisac2018safeplanning, bajcsy2020robust}. Other work considers staying safe with respect to a Gaussian noise model over the human's state \cite{liu2015safeexploration} and nonparametric models like Gaussian Processes \cite{lederer2019uniform, berkenkamp2015safe}. 
The main drawback of most methods for dealing with uncertainty in safe control is the conservatism of the resulting policy for the robot. In this work, we aim to reduce this conservatism through active exploration to improve the robot's estimated model of the human. 

\noindent\textbf{Information gathering and adaptation:} Some existing work considers a robot taking information gathering actions over a human's internal state to improve the robot's model of the human \cite{sadigh2016information,bestick2018active}. In particular, \cite{sadigh2016information} considers a model of the human as making its best response to the robot's policy, which is similar to the human model considered in this work, though we encapsulate the human's intentions inside their dynamics.

Another way to reduce the conservatism of the robot is to adapt the robot's model of the human online. Online adaptation has long been used for system identification \cite{isermann1982parameter, rohrs1985robustness} and has more recently been applied to adapting models of humans \cite{liu2020human, wang2021hierarchical}. In this work, we use active information gathering to improve the adaptation process by changing the data input to the adaptation algorithm. 

\noindent\textbf{Neural network dynamics models:} There has recently been lots of work that utilizes neural networks to represent dynamics models \cite{nagabandi2018learning,wei2021safe}. This can be particularly useful in cases where the system can be hard to model explicitly, like human behavior. The authors in \cite{chua2018deep} also consider uncertainty in the learned dynamics to attain better sample efficiency. Additionally, \cite{nagabandi2018adapt} considers adapting neural network dynamics models online through meta-learning. While there has been work on safe active learning for adaptation of neural network dynamics \cite{lew2022safe}, to the best of our knowledge, we present the first investigation of safe exploration for neural network dynamics models in human-robot interaction.

\section{Safe Exploration in HRI}
\label{sec:active_exploration}
\subsection{The HRI System}
We consider the case of a single robot interacting with a single human. The human has dynamics $\pi_{\theta}(\cdot)$ that are parameterized by the vector $\theta$ and depends on the environment state $x(k)$:
\begin{align}
    x(k) &= (x_H(k), x_R(k), x_G(k), x_G^R(k))\\
    x_H(k+1) &= \pi_{\theta}((x(k)),
\end{align}
where the environment state is defined by the human's state $x_H(k)$, the robot's state $x_R(k)$, the human's goal $x_G(k)$ and the robot's goal $x_G^R(k)$. The robot's dynamics are assumed to be control-affine (such that $f_R(\cdot)$ and $g_R(\cdot)$ are potentially nonlinear functions of $x_R(k)$); and it is equipped with a control law $\pi_R(\cdot)$ and an adaptation law to estimate the parameters $\theta$ of the human's policy $\pi_{\theta}$:
\begin{align}
    x_R(k+1) &= f_R(x_R(k)) + g_R(x_R(k))u_R(k) \label{eqn:control_affine}\\
    u_R(k) &=\pi_R(x_R(k), x_H(k), x_G^R(k), \hat\theta(k))\\
    \begin{split}
    \hat\theta(k+1) &= \text{update}(\hat\theta(k), x_H(k), x_H(k-1),\ldots,\\
    & \;\;\;\;\;\;\;\;\;\;\;\;\;\;\;\; x_R(k),x_R(k-1),\ldots)
    \end{split}
\end{align}
where $\hat\theta(k)$ is the robot's estimate of $\theta$ at time $k$. Note that the robot does not necessarily have access to the form of $\pi_{\theta}$, so $\theta$ and $\hat\theta(k)$ may not have the same semantic meaning.

\subsection{Goal for Robot Control}
The goal for robot control is to minimize the prediction error over the distribution of environment states that the human-robot system will reach when both agents are running goal-focused controllers (denoted as $\mathcal{G}$) while additionally minimizing the number of \textit{unsafe states} the system will visit. Based on a predefined safety specification (like keeping the human and robot from colliding), the system has a set of safe states $\mathcal{Z}$, so the robot incurs a penalty if the human-robot system leaves this set. The full cost function for the robot is:
\begin{equation}
    \pi_R^* = \argmin{\pi}\mathop{\mathbb{E}}\limits_{x\sim\mathcal{G}}\left[||\pi_{\theta}(x)-\hat\pi_{\hat\theta^{\pi}}(x)|| + \beta_{x\not\in \mathcal{Z}}\right]\,
    \label{eqn:objective_function}
\end{equation}
where $\hat\pi_{\hat\theta^{\pi}}(\cdot)$ denotes the robot's estimated policy of the human (parameterized by $\hat\theta^{\pi}$) after the robot executes control law $\pi$ and estimates $\hat\theta^{\pi}$ during the interaction and $\beta$ is a constant coefficient. Intuitively, minimizing this objective means that the robot should both learn a good predictive model of the human and keep the system outside of unsafe states. The objective function does not require the human's dynamics model and the robot's estimated model to have the same form since it just considers the state prediction error.

\subsection{Safety Assurance}
To minimize \eqref{eqn:objective_function}, the robot could either trade off between the two objectives or could treat the safety violation penalty as a constraint while minimizing the prediction error. We consider the latter approach where the robot has a safety monitor that keeps the system safe.

We adopt the safe exploration algorithm \cite{liu2014safeset,liu2015safeexploration} for safety assurance. The method belongs to general energy-function-based safe control \cite{wei2019safe}, which ensures forward invariance to a subset of a user-defined safe set. The safety specification considered in this paper is collision avoidance between the human and the robot. Hence the user-defined safe set $\mathcal{Z}=\{x\mid\phi_0=d_{min}-d(x_R,x_H)\leq 0\}$, where $d_{min}$ is the distance margin and $d$ measures the minimum distance between the human and the robot. The key idea of safe control is to design a safety index (or energy function) $\phi:x\mapsto \mathbb{R}$ such that 1) a control law that satisfies the constraint $\dot \phi \leq -\eta_R \text{ when } \phi\geq 0$ ensures forward invariance to the set $\mathcal{X}:=\{x\mid \phi\leq 0,\phi_0\leq 0\}$; and 2) there always exists a feasible safe control for any state $x$, i.e., $U_S^R(x) = \{u_R : \dot \phi \leq -\eta_R \text{ when } \phi\geq 0\}\neq \emptyset$. The parameter $\eta_R\in\mathbb{R}^+$ is a safety margin. Once such a safety index is constructed, we just need to construct a safety monitor to project all reference control signals to $U_S^R(x)$. Then safety is guaranteed in the sense that if the system trajectory starts from the set $\mathcal{X}$ will always remain in that set.

This work uses a second-order control-affine robot model. Hence according to the design rule in \cite{liu2014safeset}, the safety index is designed to be
\begin{equation}
    \phi = d_{min}^2+\rho -d^2 - k_{\phi}\dot d,
    \label{eqn:safety_index}
\end{equation}
where $\dot d$ is the time derivative of the relative distance, $k_{\phi}$ is a positive constant, and $\rho$ is a positve margin. The derivative term $\dot d$ is added in order to ensure that the robot control $u_R$ can always affect $\dot\phi$. Suppose the next human state follows a Gaussian distribution $x_H(k+1)\sim \mathcal{N}(\hat x_H(k+1\mid k),\Sigma_H(k+1))$ (to be derived in \sref{sec:math_background}), then
\begin{align}
    \dot\phi(x(k)) \sim  & \mathcal{N}(\frac{\partial\phi}{\partial x_R}\dot x_R + \frac{\partial\phi}{\partial x_R}\hat{\dot x}_H,\nonumber\\
    & \frac{1}{t_s} \left[ \left(\p{\phi}{x_H}\right) \Sigma_H(k+1) \left(\p{\phi}{x_H}\right)^T \right]^{\frac{1}{2}}),
\end{align}
where $t_s$ is the sampling time, $\dot x_R = \nicefrac{(x_R(k+1)-x_R(k))}{t_s}$, and $\hat{\dot x}_H:=\nicefrac{(\hat x_H(k+1\mid k)-x_H(k))}{t_s}$. 
Since the distribution of $\dot\phi$ is unbounded, we can only enforce a probabilistic constraint on $\dot\phi$ instead of a hard constraint. This paper aims to ensure safety for human behaviors within the $3\sigma$ bound. In this way, the $3\sigma$-robust set of safe control $U_S^R(x) = \{u_R : \mathcal{P}(\dot \phi \leq -\eta_R)\geq 99.7\% \text{ when } \phi\geq 0\}$ at time $k$ can be computed as described in \cite{liu2015safeexploration}:
\begin{align}
    U_S^R &= \{u_R(k) : L(k) u_R(k) \leq S(k)\}\label{eqn:set_of_safe_control}\\
    L(k) &= \p{\phi}{x_R}g_R \\
    S(k) &= \begin{cases}
                -\eta_R - \lambda^{SEA}_R(k) - \p{\phi}{x_H}\hat{\dot x}_H - \p{\phi}{x_R}f_R & \text{if $\phi \geq 0$}\\
                \infty & \text{if $\phi < 0$}\\
            \end{cases}
    \label{eqn:safety_const}
\end{align}
where 
\begin{equation}
    \lambda^{SEA}_R(k) = \frac{3}{t_s} \left[ \left(\p{\phi}{x_H}\right) \Sigma_H(k+1) \left(\p{\phi}{x_H}\right)^T \right]^{\frac{1}{2}} + \lambda_0.
    \label{eqn:lambda_SEA}
\end{equation}
Here, $\lambda_0$ is a tunable constant that bounds other uncertainties. In practice, in order to compensate discrete-time implementation error in safe control, we choose the margin $\rho$ as $\eta_R t_s + \lambda^{SEA}_R(k) t_s$, which grows with the uncertainty of $\dot\phi$ (that is affected by the uncertainty of human behaviors). 

Finally, our safety controller can be written as the solution to a quadratic program that maps some reference control $u_R^{ref}(k)$ (e.g., actions for exploration) to the set of safe controls $U_R^S$ as the control in the safe set that is closest to the reference:
\begin{equation}
    u_R^* = \argmin{u_R\in U_R^S} \frac{1}{2}(u_R-u_R^{ref})^T(u_R-u_R^{ref}).
\end{equation}

When there is no uncertainty on human behavior, i.e., $\Sigma_H \equiv 0$, the safe control enforces forward invariance to $\mathcal{X}$. If there is uncertainty and the uncertainty model is correct, i.e., $\Sigma_H$ matches the statistical error covariance in the human behavior prediction, then the safe control enforces forward invariance with at least probability 99.7\%.

\subsection{Active Exploration}
The robot's policy will affect \eqref{eqn:objective_function} by changing the data input to the adaptation, which will in turn change its estimate of the human's policy $\hat\pi_{\hat\theta^{\pi}}(\cdot)$. 
The human's policy could be estimated passively by just receiving information during the interaction, or actively by choosing actions ($u_R^{ref}$) to reduce the future error. Active exploration may minimize the immediate or long-term error, which fall on the spectrum of different risk preferences, where ``risk'' refers to the chance of incorrectly predicting the human's state:

\noindent\textbf{Risk-Neutral:} the robot does not explicitly reason about uncertainty and adapts its model passively.\\
\noindent\textbf{Risk-Seeking:} the robot tries to actively explore unknown states with high uncertainty, with the hope to reduce uncertainty in the subsequent rounds.\\
\noindent\textbf{Risk-Averse:} the robot actively keeps the system in known states with low uncertainty.


\subsection{Types of Uncertainties}

We're interested in understanding the effects of these risk preferences on the adaptation process while the robot acts with its safety monitor to understand \textit{when safe active exploration can improve the robot's model of the human}.
We investigate the effect of active exploration when the robot's human model has different kinds of uncertainty:

\noindent\textbf{Intrinsic Uncertainty:} The robot is estimating the portion of the human's dynamics that \textit{does not} depend on the robot's state, so is uncertain of how the human moves in the environment.\\
\noindent\textbf{Interactive Uncertainty:} The robot is estimating the portion of the human's dynamics that \textit{depends on} the robot's state, so is uncertain of how the human reacts to the robot.\\
\noindent\textbf{Full Uncertainty:} The robot estimates both the human's intrinsic and interactive dynamics, so is uncertain about how the human moves in the environment and how it reacts to the robot.


\section{Online Adaptation of Human Models}
\label{sec:math_background}


In the standard formulation \cite{ljung1987theory}, the robot will estimate a discrete-time parameter-affine system as the human's dynamics model:
\begin{equation}
    x_H(k+1) = \Phi(k)\theta(k) + w_H(k),
    \label{eqn:human_dyn}
\end{equation}
where the matrix $\Phi(k)$ is some observation of the environment (e.g. nonlinear features of the human's state and control), $\theta(k)$ is the parameter to be estimated and $w_H(k)$ is assumed to be zero-mean Gaussian noise with covariance $W$. In the following discussion, we introduce a linear parameterization of the human model and a recursive least square parameter adaptation (RLS-PA) algorithm to identify unknown parameters. Notation for the RLS-PA is shown in Table \ref{tab:paa_definitions}.

\subsection{Time-Varying Linear Model}
We follow prior work by regarding the model of the human as a time-varying linear system with Gaussian uncertainty \cite{liu2015safeexploration}, meaning the robot estimates time-varying parameters $A_H$ and $B_H$:
\begin{equation}
    x_H(k+1) = A_H(k) x_H(k) + B_H(k) u_H(k).
    \label{eqn:linear_system}
\end{equation}
We design a feature vector $u_H$ in place of the human's true control vector, since the robot does not have access to it. $u_H$ could generally include arbitrary features, but we assume it's a function of the human's state $x_H(k)$, the robot's state $x_R(k)$ and the human's goal $x_G(k)$:
\begin{equation}
    u_H(k) = g_H(x_H(k), x_R(k), x_G(k)).
    \label{eqn:u_H}
\end{equation}

At time $k+1$, the robot uses its previous estimate of the state and dynamics to get the \textit{a priori} state estimate:
\begin{equation}
    \hat{x}_H(k+1|k) = \hat{A}_H(k)\hat{x}_H(k|k) + \hat{B}_H(k) u_H(k).
\end{equation}
The state estimation error is defined as 
\begin{equation}
    \tilde{x}_H(k+1|k) = x_H(k+1) - \hat{x}_H(k+1|k).
\end{equation}
For simplicity, we assume that the robot can access the ground truth human state, hence the \textit{a posteriori} state estimate has no error, i.e., $\hat x_H(k|k) = x_H(k)$ and $\tilde x_H(k|k) = 0$.
\begin{table}
    \centering
    \begin{tabular}{@{}llll@{}}
    \toprule
                          & State Estimate    & Estimation Error    & State Covariance \\ \midrule
    \textit{a priori}     & $\hat x_H(k|k)$   & $\tilde x_H(k|k)$   &                  \\ \midrule
    \textit{a posteriori} & $\hat x_H(k+1|k)$ & $\tilde x_H(k+1|k)$ & $\Sigma_H(k+1)$  \\ \bottomrule
    \end{tabular}
    \caption{\label{tab:paa_definitions} Notation for State Estimation}
    \vspace{-0.2in}
\end{table}

\subsection{State and Parameter Estimation in the Belief Space}
\label{ssec:belief_space}
The robot keeps a Gaussian uncertainty model on its estimate of the human
\begin{equation}
    \hat x_H(k+1) \sim \mathcal{N}(\hat x_H(k+1|k), \Sigma_H(k+1)),
    \label{eqn:gaussian}
\end{equation}
where $\hat x_H(k+1|k)$ is the \textit{a priori} estimate of the human's next state and $\Sigma_H(k+1)$ is the state covariance. 

To put the human model into a parameter-affine form \eqref{eqn:human_dyn}, we define the matrix $C = \begin{bmatrix}A_H(k) & B_H(k)\end{bmatrix}$, and then split it into its rows to consider the flattened $\theta = \begin{bmatrix}C_1 & \dots & C_{n_h}\end{bmatrix}^T$ where $n_h$ is the dimension of the human state and $C_{i}$ denotes the $i$th row of $C$. The robot's estimate of $\theta$ is $\hat\theta(k)$. Next, the observation matrix $\Phi(k)$ is
\begin{equation}
    \Phi(k) = \begin{bmatrix}\varphi^T(k) & 0 & \dots & 0\\
                            0 & \varphi^T(k) & \dots & 0\\
                            \vdots & \vdots & \ddots & \vdots\\
                            0 & 0 & \dots & \varphi^T(k)
    \end{bmatrix}
    \label{eqn:Phi}
\end{equation}
where $\varphi(k)$ is the observation vector 
\begin{equation}
    \varphi(k) = \begin{bmatrix}\hat x_H(k|k)\\ u_H(k)\end{bmatrix}.
    \label{eqn:varphi}
\end{equation}
This allows us to write our \textit{a priori} state estimate simply as
\begin{equation}
    \hat x_H(k+1|k) = \Phi(k)\hat\theta(k)
\end{equation}
The parameter estimation error is $\tilde\theta(k) = \theta(k) - \hat\theta(k)$. This lets us express the state estimation error as
\begin{equation}
    \tilde x_H(k+1|k) = \Phi(k)\tilde\theta(k) + w_H(k)
    \label{eqn:x_tilde}
\end{equation}
Now we consider the state covariance, or mean squared estimation error $\Sigma_H(k+1) = \mathbb{E}[\tilde x_H(k+1|k) \tilde x_H(k+1|k)^T]$. Using \eqref{eqn:x_tilde}, we can rewrite this as 
\begin{equation}
    \Sigma_H(k+1) = \Phi(k)\Sigma_{\tilde \theta\tilde\theta}(k)\Phi^T(k) + W
\end{equation}
where $W$ is the (known) measurement noise covariance and $\Sigma_{\tilde \theta\tilde\theta}(k) = \mathbb{E}[\tilde\theta(k)\tilde\theta(k)^T]$ is the covariance of the model error. 

Now, we'll consider how the robot adapts its model of the human in the belief space: 
\begin{equation}
    \hat\theta(k+1) = \hat\theta(k) + F(k+1)\Phi^T(k)\tilde x_H(k+1|k)
\end{equation}
where $F(k+1)$ is the learning gain which is updated as:
\begin{equation}
    \begin{split}
        &F(k+1) = \frac{1}{\lambda}[F(k) - \\&F(k)\Phi^T(k)(\lambda I + \Phi(k) F(k) \Phi^T(k))^{-1}\Phi(k) F(k)].
    \end{split}
    \label{eqn:learning_gain}
\end{equation}

The parameter estimation error is 
\begin{equation}
    \tilde\theta(k+1) = \tilde\theta(k) - F(k+1)\Phi^T(k)\tilde x_H(k+1|k) + \Delta\theta
\end{equation}
where $\Delta\theta=\theta(k+1) - \theta(k)$. The true value of $\tilde\theta(k)$ is unknown, but we can calculate its expectation as 
\begin{equation}
        \mathbb{E}[\tilde\theta(k+1)] =[I - F(k+1)\Phi(k)\Phi^T(k)]\mathbb{E}[\tilde\theta(k)] + d\theta
\end{equation}
where $d\theta$ is set to an average time varying rate, since the true $\Delta\theta$ is also unknown.

Finally, this lets us write an explicit update for our model parameter covariance as: 
\begin{equation}
    \begin{split}
        \Sigma_{\tilde \theta\tilde\theta}(k+1) &= F(k+1)\Phi(k)\Sigma_H(k+1)\Phi(k)F(k+1) \\
        &- \Sigma_{\tilde \theta\tilde\theta}(k)\Phi^T(k)\Phi(k)F(k+1) \\
        &- F(k+1)\Phi^T(k)\Phi(k)\Sigma_{\tilde \theta\tilde\theta}(k)\\
        &+ \mathbb{E}[\tilde\theta(k+1)]d\theta^T + d\theta\mathbb{E}[\tilde\theta(k+1)]^T\\
        &- d\theta d\theta^T + \Sigma_{\tilde \theta\tilde\theta}(k)
    \end{split}
    \label{eqn:msee_theta}
\end{equation}
 
\section{Exploration Strategies}

We now substantiate the exploration strategies that we introduced earlier using the estimates from the online adaptation algorithm. 
The robot's future uncertainty can be measured by looking at the norm of the parameter uncertainty matrix $||\Sigma_{\tilde\theta\tilde\theta}(k+1)||$. However, since $\Sigma_{\tilde\theta\tilde\theta}(k+1)$ does not depend on the robot's action at time $k$, we need to optimize for the uncertainty two steps in the future: $\Sigma_{\tilde\theta\tilde\theta}(k+2)$. Note that the baseline risk-neutral controller is the same controller used in \cite{liu2015safeexploration}.

\label{ssec:risk_cost_functions}
\noindent\textbf{Risk-Neutral:} Risk-neutral behavior will not account for uncertainty and just move the robot towards its own goal with state feedback control gain $K$ (hand-tuned):
\begin{equation}
    u_R^{ref}(k) = -K(x_R(k) - x_G^R(k)).
\end{equation}
\noindent\textbf{Risk-Seeking:} Risk-seeking behavior will try to maximize the objective function $J(\cdot)$, which is the norm of the future covariance matrix:
\begin{equation}
    u_R^{ref}(k) = \argmax{u_R} J(u_R) = ||\Sigma_{\tilde\theta\tilde\theta}(k+2)||.
    \label{eqn:risk_seeking}
\end{equation}
\noindent\textbf{Risk-Averse:} Risk-averse behavior will try to minimize $J(\cdot)$, the norm of the future covariance matrix: 
\begin{equation}
    u_R^{ref}(k) = \argmin{u_R} J(u_R) = ||\Sigma_{\tilde\theta\tilde\theta}(k+2)||.
    \label{eqn:risk_averse}
\end{equation}

The types of uncertainties in $\theta$ that need to be explored are summarized below:
\label{ssec:uncertainty_types}

\noindent\textbf{Intrinsic Uncertainty:} The robot estimates only $A_H$.\\
\noindent\textbf{Interactive Uncertainty:} The robot estimates only $B_H$.\\
\noindent\textbf{Full Uncertainty:} The robot estimates both $A_H$ and $B_H$.

\section{Analytical Human Model Experiments}
\label{sec:simulations}
\subsection{Analytical Human Model}
The robot's state $x_R(k)$ and the human's state $x_H(k)$ consist of each agent's own position and velocity. The goal state for the robot $x_G^R(k)$ and the goal state for the human $x_G(k)$ are defined as points in the state space with $0$ velocity. We hand-design features for the simulated human's control $u_H(k)$ \eqref{eqn:u_H}. Since the human's objective in the task is to move towards their goal while avoiding collisions with the robot, we consider a potential field model for the human's control---the human is attracted to the goal and repelled from the robot:
\begin{equation}
    u_H(k) = -K_1\left(x_H(k)-x_G(k)\right) + \frac{\gamma}{d^2}K_2\left(x_H(k)-x_R(k)\right),
    \label{eqn:u_H_sim}
\end{equation}
where $K_1$ and $K_2$ are constant gains and $d$ is the distance between the two agents. The second component of the control vector captures the idea that the robot's influence on the human will decrease as the distance between the two increases and this influence is controlled by the parameter $\gamma$. In the evaluation (unless otherwise noted), $\gamma=30$.
\begin{figure}[t!]
    \centering
    \includegraphics[width=\columnwidth]{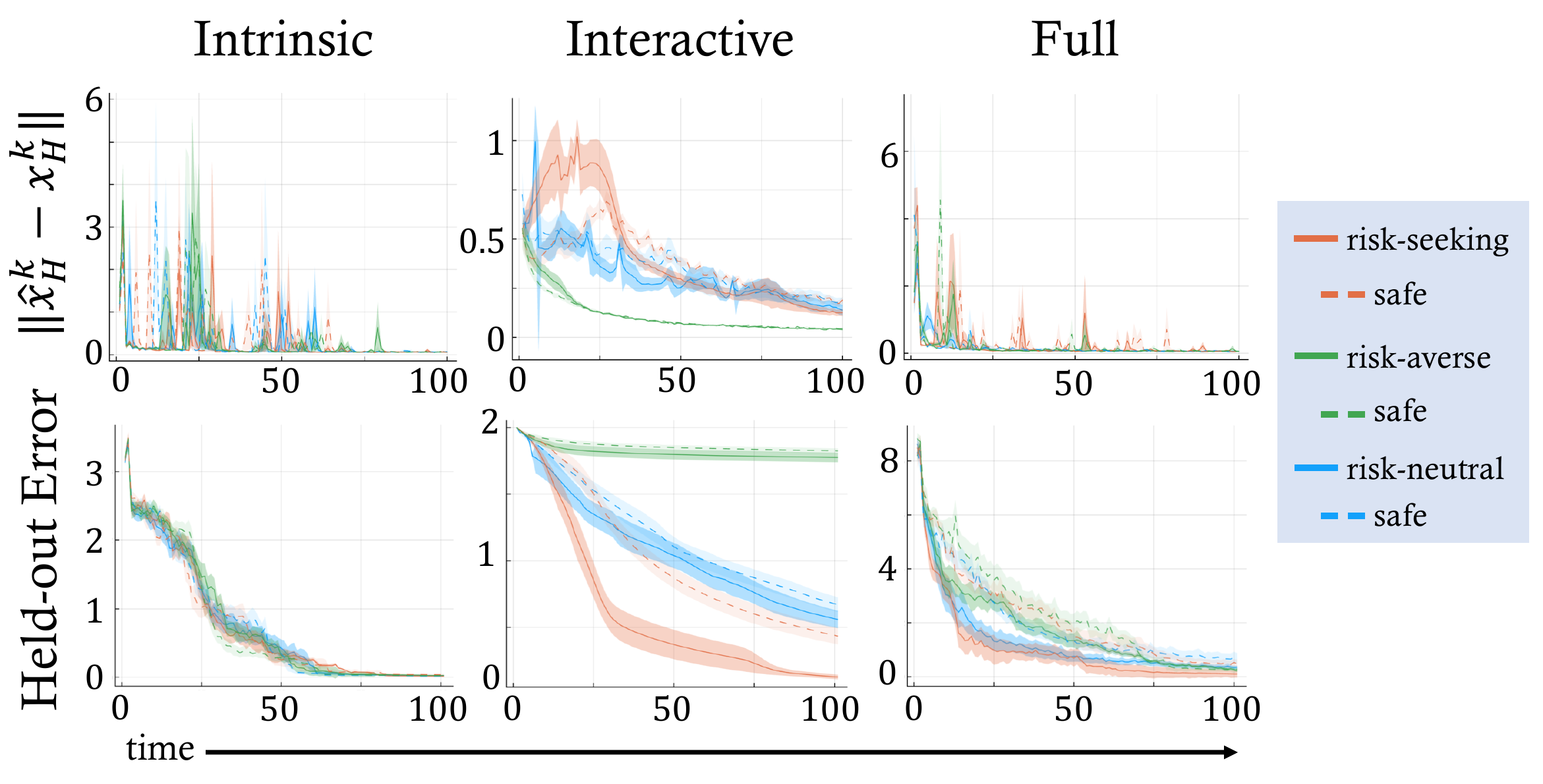}
    \caption{Each column corresponds to simulations with different kinds of uncertainty (\sref{ssec:uncertainty_types}). Each row corresponds to a different metric (\sref{ssec:evaluation}). Each curve is averaged over 10 different initial conditions and the shaded areas show the standard error.}
    \label{fig:avg_metrics}
\end{figure}

\subsection{Hypotheses}
\noindent\textbf{H1:} The risk-seeking safe exploration strategy will improve the quality of the robot's estimated model of the human.

\noindent\textbf{H2:} The presence of the safety controller will decrease the efficiency of adaptation.

\noindent\textbf{H3:} Active exploration will be most beneficial when the robot has interactive uncertainty in the human's dynamics.

\begin{table}
    \centering
    \begin{tabular}{@{}llll@{}}
    \toprule
    & risk-seeking   & risk-averse   & risk-neutral  \\ 
    \midrule intrinsic uncertainty & $4.6 \pm 0.2$ & $4.6 \pm 0.3$ & $5.2 \pm 1.7$ \\ \midrule
    interactive uncertainty & $19.7 \pm 2.3$  & $0.9 \pm 0.09$ & $6.3 \pm 2.0$ \\ \midrule
    full uncertainty & $13.2 \pm 3.5$  & $4.5 \pm 0.7$ & $7.0 \pm 1.4$ \\ \midrule
    neural network dynamics & $7.2 \pm 1.3$  & $4.6 \pm 2.5$ & $4.2 \pm 1.8$ \\ \bottomrule
    \end{tabular}
    \caption{\label{tab:safety_violations} Number of Safety Interventions (mean $\pm$ SD)}
\end{table}

\subsection{Evaluation}
\label{ssec:evaluation}
In our simulated navigation environment, the robot needs to adapt its model of the human because its initial guesses for $A_H$ and $B_H$ are incorrect. We care about three metrics to measure the quality of the adapted model and the effect of the safety controller:

\noindent\textbf{Runtime Error:} Ultimately, we want the robot to end up with a better model of the human based on the actions it selects, so we measure this with the 1-step prediction accuracy of the model on the current trajectory.

\noindent\textbf{Held-out Error:} To understand if the robot's estimated model generalizes, we measure the first term of the objective function \eqref{eqn:objective_function} by computing the average 1-step state prediction error of the estimated model on rollouts of a set of different initial conditions (without further adaptation) where the robot is only taking goal-oriented actions.

\noindent\textbf{Safety Interventions:} When the robot is getting close to violating the minimum safe distance constraint, the safety controller will activate to keep the robot safe, so we can measure how often this happens to understand how exploration is affected by the presence of the the safety controller.

\begin{figure}[t!]
    \centering
    \includegraphics[width=\columnwidth]{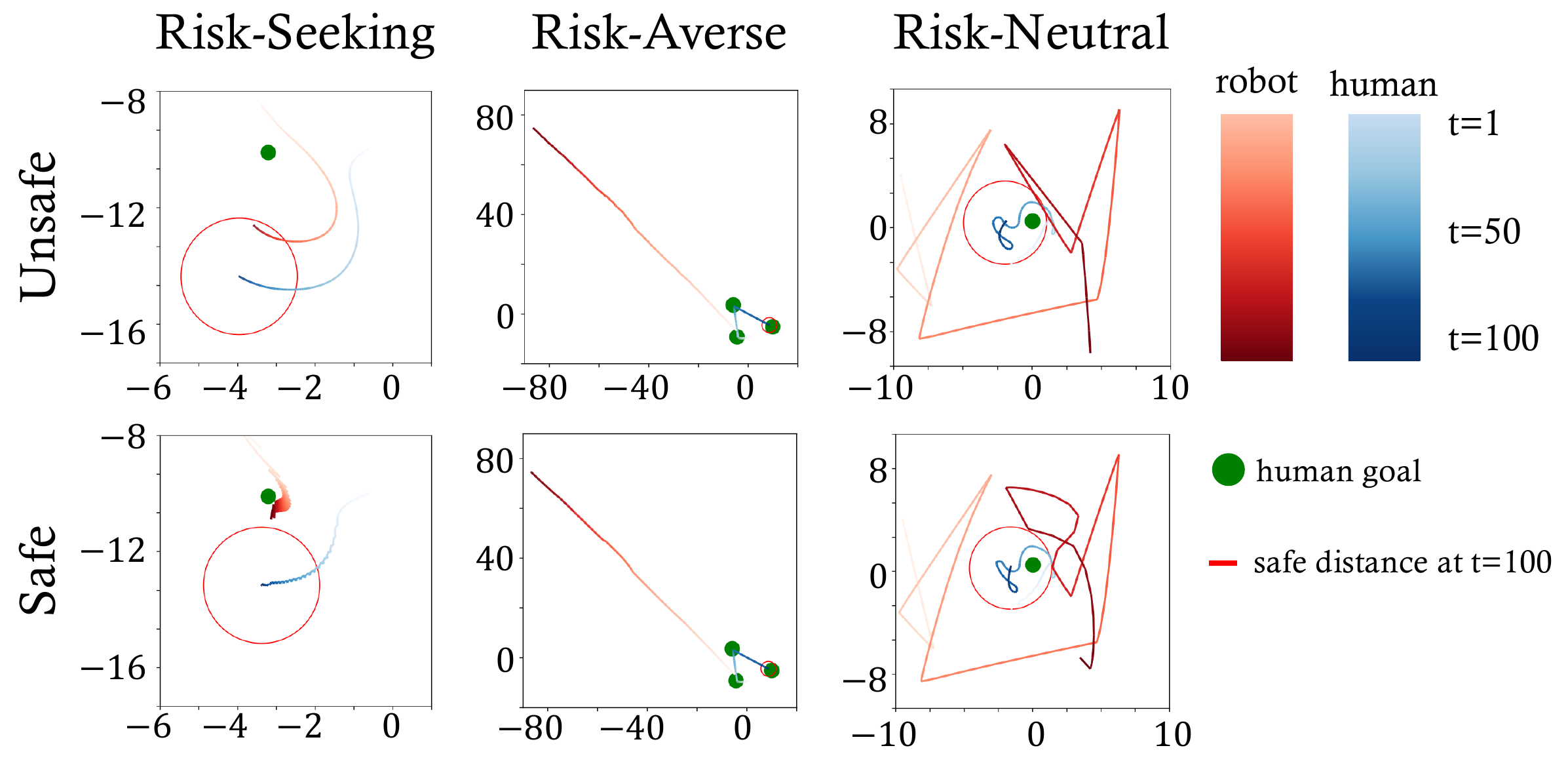}
    \caption{One example trajectory for each risk preference when the robot has interactive uncertainty (\sref{ssec:interactive}). \textbf{Top row:} without safety controller \textbf{Bottom row:} with safety controller.}
    \label{fig:trajectories}
    \vspace{-0.2in}
\end{figure}

\subsection{Intrinsic Uncertainty}
\label{ssec:intrinsic}
Each column in \figref{fig:avg_metrics} shows results for a different kind of uncertainty and each row shows a different metric. Each curve on a single plot shows the value of the metric over a 100-timestep interaction between the human and robot, averaged over 10 initial conditions (randomly selected starting locations and goals in the xy-plane for both agents). Safety interventions are shown in \tabref{tab:safety_violations}.

When the robot has only intrinsic uncertainty, the performance of all risk preferences with and without safety are basically the same, shown in the left column of \figref{fig:avg_metrics} (the spikes in prediction error occur when the human's goal changes). This result gives evidence against \textbf{H1} and \textbf{H2}, since neither active exploration nor the safety controller had any effect on the robot's adaptation process. However, this makes sense because the robot's model of the human is not a function of the robot's state, so the robot is able to estimate the human's intrinsic dynamics well regardless of the robot's trajectory. This result also supports \textbf{H3}, since neither risk-seeking nor risk-averse behavior improved the robot's adapted model of the human when the robot has intrinsic uncertainty.

The top row of \tabref{tab:safety_violations} shows that the number of interventions from the safety controller is similar for all three risk preferences, telling us that the safety controller affected the different risk preferences similarly.

\subsection{Interactive Uncertainty}
\label{ssec:interactive}
When the robot has only interactive uncertainty, we see significant differences between the different risk preferences, shown in the middle column of \figref{fig:avg_metrics}. The risk-averse controller converges to low runtime error quickly while the other risk preferences eventually also reach similarly low values. This happens because the risk-averse controller moves away from the human (\figref{fig:trajectories}), so it will barely affect the human's trajectory, but since it already knows $A_H$, it can easily predict the future states of the human.

We did not see this pattern when the robot had only intrinsic uncertainty, so this result also supports \textbf{H3}. The risk-seeking controller results in much lower held-out error than the baseline risk-neutral controller, which supports \textbf{H1} and \textbf{H3}. The risk-averse controller, however, results in worse held-out error even though its prediction error is low. We can again understand why by looking at the middle column of \figref{fig:trajectories}---the risk-averse controller runs away from the human to keep its uncertainty low, so it does not get to learn the effect it has on the human and thus its estimated model does not generalize well. 

Comparing safe exploration (dashed curves) to unsafe exploration (solid curves), we see that the safety controller does slightly increase the held-out model error for all risk preferences, supporting \textbf{H2}, though this difference is most pronounced for risk-seeking controller. We can understand why by looking at \figref{fig:trajectories}, which shows one example interaction for each risk preference with and without the safety controller. The safety controller changes the risk-seeking controller's trajectory drastically, but not the other two risk preferences' trajectories. This is a direct result of the risk-seeking controller trying to stay close to the human, so the safety controller will need to be active more often.

We quantitatively see this in the second row of \tabref{tab:safety_violations}---the safety controller was activated an average of 19.7 times during a 100-timestep interaction for the risk-seeking controller, while it was activated less than once on average for the risk-averse controller and 6.3 times for risk-neutral. This confirms that the risk-seeking controller was more affected by the presence of the safety controller than the other risk preferences were, showing that \textbf{H2} being true depends on both the kind of uncertainty and risk preference considered.

\subsection{Full Uncertainty}
\label{ssec:full}
In this case, the robot's dynamics estimate has both intrinsic and interactive uncertainty. The results in these simulations (right column of \figref{fig:avg_metrics}) show the same trends for held-out error as in the interactive uncertainty case, which makes sense because this model also includes interactive uncertainty, so this again supports \textbf{H3}. We also see that the risk-seeking controller results in the lowest held-out model error, supporting \textbf{H1}. 

Introducing the safety controller makes the held-out error higher for all three risk preferences, which lines up with \textbf{H2} just as in the interactive uncertainty case. We again see the same pattern that the risk-seeking controller's model gets affected the most, which can be explained by considering the number of safety interventions (third row of \tabref{tab:safety_violations})---the risk-seeking controller has the most number of interventions, so its trajectory is affected most by the safety controller.

\begin{figure}[t!]
    \centering
    \includegraphics[width=\columnwidth]{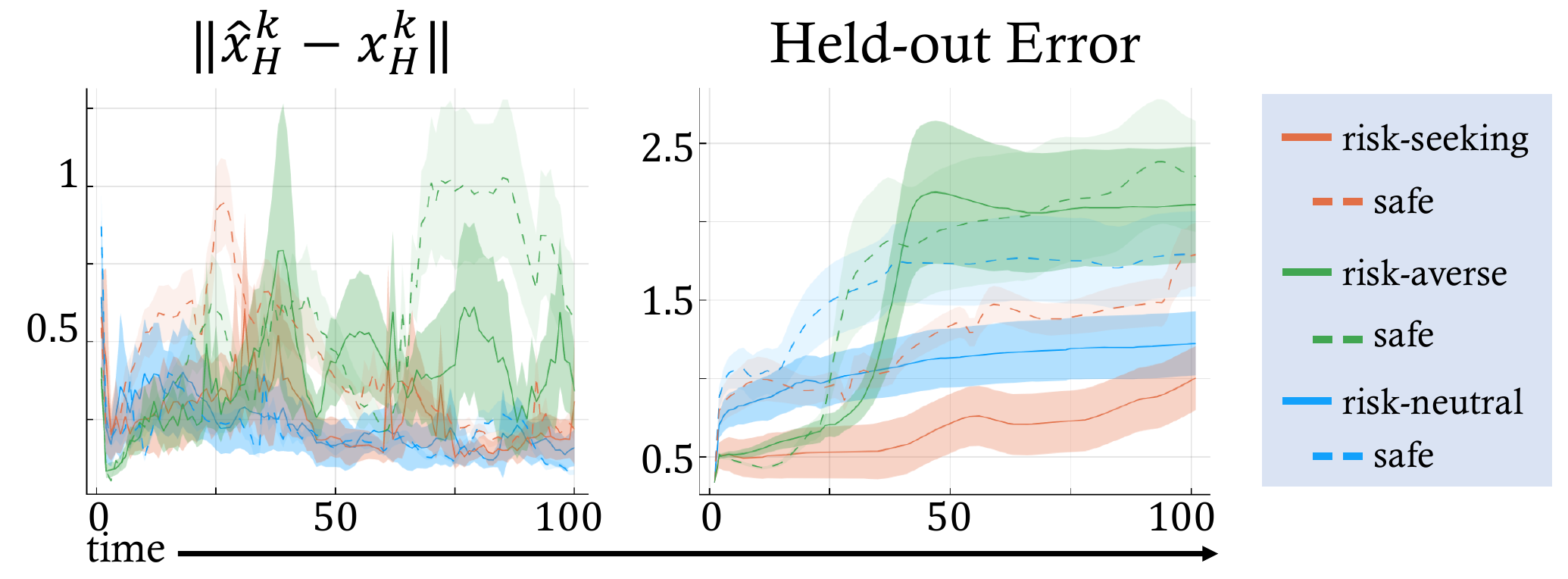}
    \caption{Each column corresponds to a different metric (\sref{ssec:evaluation}) in the case when the robot is estimating a neural network to represent the human's dynamics. Each curve is averaged over 10 different initial conditions and the shaded areas represent the standard error of the mean.}
    \label{fig:metrics_learned}
    \vspace{-0.2in}
\end{figure}

\section{Neural Network Human Model Experiments}
\label{sec:learning_human}

Human behavior is diverse and may not always be captured well by an analytical model. To capture the diversity of human behavior, data-driven neural network models are often used. In the following discussion, we test the same risk preferences and safe controller while the robot keeps a neural network estimate of the human's dynamics.

\subsection{Dataset and Architecture}
\label{ssec:nn_dataset}
We created a dataset collected from the human model in \eqref{eqn:u_H_sim}. The dataset $\mathcal{D}$ consists of trajectories of length $T=100$ of the human's state, the human's goal and robot's state $\tau = (x_H^0,x_R^0,x_G^0,...,x_H^T,x_R^T,x_G^T)$. We then split these trajectories into segments with a short history of length $N$ to use as training inputs $(x_H^{k-N},x_R^{k-N},x_G^{k-N},...,x_H^k,x_R^k,x_G^k)$ and their corresponding labels $x_H^{k+1}$. 

We train a 4-layer feedforward ReLU neural network to learn the human's dynamics function. The network is trained by minimizing an MSE loss with the Adam optimizer.



\subsection{Last Layer Adaptation}
Since the robot gets to train a good model offline in this case, we make adaptation necessary by generating the training data with $\gamma=50$ but the human's true value is $\gamma=30$ online. 

To adapt the model online, we can adapt the last layer's weights (a common paradigm for fine-tuning neural network models \cite{yosinski2014transferable}). Using the flattened post-activation output of the second to last layer of the network as $\varphi(k)$ and the weights of the last layer to be $\hat\theta(k)$, we can keep the same form of the dynamics as in \eqref{eqn:human_dyn}. We adapt $\hat\theta(k)$ online and keep track of our parameter uncertainty $\Sigma_{\tilde\theta\tilde\theta}(k)$ as before. 

\subsection{Results}
Unlike in the linear system case, we can't easily separate the robot's uncertainty into intrinsic and interactive uncertainty, so the robot has full uncertainty by being uncertain about the parameters in the last layer of the network. 

Looking at \figref{fig:metrics_learned}, the prediction error does not show any strong patterns in with either the risk preferences or the safety controller, besides that the the risk-averse controller has slightly higher error (this is likely due to the robot's state going far outside the range of values in the training data for the neural network when it runs away from the human).

In the held-out error, we see a similar pattern as in the linear system case emerge with the risk preferences and the safety controller. The risk-seeking controller results in the best learned model while the risk-averse controller has the worst, which supports \textbf{H1}. The held-out error does slightly increase over time, which is not unexpected since the goal of adaptation is to overfit to temporally local data, so it may ``forget'' its previous training. This effect doesn't appear in the linear system case because improving local prediction accuracy will also improve global accuracy.

For the risk-seeking and risk-averse controllers, the presence of the safety controller reduces the quality of the learned model, which again supports \textbf{H2}. The decrease in model quality is again higher for the risk-seeking controller than the others, which is likely a result of the same phenomenon observed in the linear system. We can again see evidence for this in the bottom row of \tabref{tab:safety_violations} where the safety controller was activated most often for the risk-seeking controller.

\section{Effect of Influence}
\begin{figure}[t!]
    \centering
    \includegraphics[width=\columnwidth]{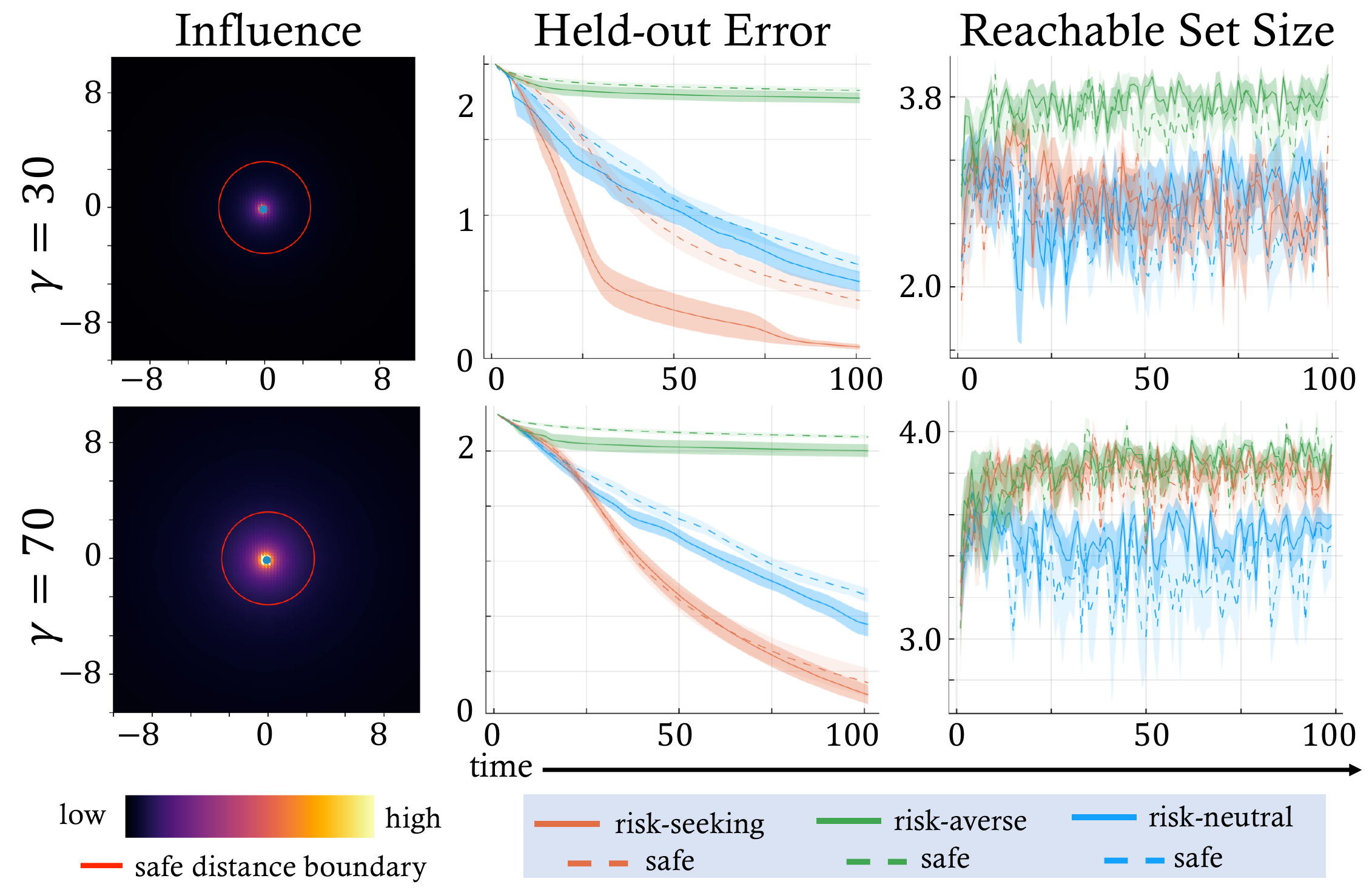}
    \caption{Shows the held-out error and reachable set size (\sref{ssec:reachable_set}) for different levels of influence (the color represents how far the human would move at the next timestep if the robot was at that point in space) when the robot has interactive uncertainty.}
    \label{fig:influence}
    \vspace{-0.25in}
\end{figure} 
\subsection{Effect on Held-out Error}
We saw previously that under interactive uncertainty, the safety controller significantly changes the reference control from the risk-seeking controller, meaning the robot has to stay farther away from the human. This results in the robot's estimate of the human's dynamics being worse than it would be without the safety controller active since observing these close-proximity states is likely ideal for adaptation---this is exactly the underlying tension between optimizing for safety and optimizing for efficiency.

To test this idea, we vary $\gamma$ in \eqref{eqn:u_H_sim} when the robot has interactive uncertainty and a linear model, since this is where we find the most stark differences between controllers. Each row of \figref{fig:influence} visualizes a different value of $\gamma$. The left column visualizes the influence that the robot has on the human where the human is in the center, the red circle shows the minimum safe distance, and the color represents how far the human would move at the next timestep if the robot was at that point in space. When $\gamma=30$, the influence is close to $0$ when the robot is outside the safe distance boundary, whereas the robot still has a notable influence on the human outside this bubble when $\gamma=70$. 

When $\gamma=70$, the safety controller \textit{does not} significantly affect the held-out error under any risk preference. This is in contrast to the $\gamma=30$ case where introducing the safety controller significantly reduced the quality of the estimated model for the risk-seeking controller. This means \textbf{H1} is still true with safe control and that \textbf{H2} is not necessarily true, it depends on how much influence the robot has on the human. This result tells us that if the human is sufficiently affected by the robot's state outside of the safe bubble, the robot can actually make use of active exploration to improve the robot's model of the human while staying safe.

\subsection{Effect on Reachable Set of States}
\label{ssec:reachable_set}
Finally, we want to understand how safe exploration can reduce the conservatism of the safety monitor, so we show a preliminary analysis here. The right column of \figref{fig:influence} shows the average size of the 1-step reachable states under safe control from the current state ($x_0$), i.e, $|\{x\mid\exists u\in U_S^R \text{, s.t., } x = f_R(x_0)+g_R(x_0)u\}|$. This set measures the conservatism of the constraint $U_S^R$ in \eqref{eqn:set_of_safe_control} under the current human model and uncertainty. Computation-wise, it is calculated by sampling controls uniformly between the control bounds, computing the resulting safe states using $\Sigma_H(k)$, then again sampling the state space to estimate the volume of the resulting safe reachable set of states. The curves show how the size of this set changes during the interaction. 

When the influence is low, the risk-averse controller keeps the safe set the largest while the risk-seeking and risk-neutral controllers are indistinguishable. When the influence is high, however, both active exploration controllers (risk-seeking and risk-averse) keep the safe reachable set larger than risk-neutral behavior, but for different reasons. The risk-seeking controller learns a good predictive model of the human, so its reachable set will enlarge while it stays close to the human, while the risk-averse controller enlarges this set by staying far away from the human (\figref{fig:trajectories}).

\section{Conclusion}
\subsection{Summary}
We have looked into the effect of introducing active exploration for adaptation in an energy-function-based safe control framework. We investigated the effects of different risk preferences (risk-seeking, risk-neutral and risk-averse) on different kinds of uncertainty (intrinsic and interactive) both on an analytical and neural network model of a human partner. The risk-seeking controller generally learns the best model of the human when there is interactive uncertainty present, though active exploration does not change the adapted model when there is only intrinsic uncertainty. Safe exploration can improve the robot's model of the human and as a result reduce the conservatism of the safety monitor.

We have also seen that the benefit of using a risk-seeking controller can be smaller when a safety controller is active since the risk-seeking controller tries to stay too close to the human. However, if the human is sufficiently influenced by robot's position, this difference can disappear. Broadly, this is a good sign for future work involving physical humans and robots, since it means we can enable robots to explore safely without negatively impacting the adaptation process, assuming the human is sufficiently influenced by the robot.

\subsection{Limitations and Future Work}
While our work focuses on the effects of different safe exploration strategies in human-robot interaction, this work is in simulation without real humans in the loop. Future work will focus on speeding up the computation of the risk-seeking and risk-averse controllers to solve them in real-time around humans. Extending this work to physical situations like collaborative manufacturing is an exciting future direction since it would require the human and robot to stay in close proximity to each other while staying safe, so having a good model of the human is paramount to the team's success.

\section*{Acknowledgments}
This material is based upon work supported by the National Science Foundation Graduate Research Fellowship under Grant No. DGE1745016 and DGE2140739 and additionally under Grant No. 2144489. Any opinions, findings, and conclusions or recommendations expressed in this material are those of the author(s) and do not necessarily reflect the views of the National Science Foundation.

\bibliographystyle{IEEEtran}
\bibliography{references}

\end{document}